\documentclass[sigconf, natbib=true]{acmart}
\AtBeginDocument{%
  \providecommand\BibTeX{{%
    \normalfont B\kern-0.5em{\scshape i\kern-0.25em b}\kern-0.8em\TeX}}}

\setcopyright{acmcopyright}
\copyrightyear{2018}
\acmYear{2018}
\acmDOI{XXXXXXX.XXXXXXX}

\acmConference[Conference acronym 'XX]{Make sure to enter the correct
  conference title from your rights confirmation emai}{June 03--05,
  2018}{Woodstock, NY}
\acmPrice{15.00}
\acmISBN{978-1-4503-XXXX-X/18/06}
\usepackage{footmisc}
\usepackage{graphicx}
\usepackage{diagbox}
\usepackage{multirow}
\usepackage{array}
\usepackage{enumitem}

\begin{document}

\title{Graph Meets LLM: A Novel Approach to Collaborative Filtering
for Robust Conversational Understanding}

\author{Zheng Chen}
\authornote{Authors contributed equally to this research. Authors alphabetically ordered by last name.}
\email{zgchen@amazon.com}
\orcid{0000-0003-4406-2193}
\affiliation{%
  \institution{Amazon Alexa AI}
  \city{Seattle}
  \state{WA}
  \country{USA}
}

\author{Ziyan Jiang}
\authornotemark[1]
\email{ziyjiang@amazon.com}
\affiliation{%
  \institution{Amazon Alexa AI}
  \city{Seattle}
  \state{WA}
  \country{USA}
}

\author{Fan Yang}
\authornotemark[1]
\email{ffanyang@amazon.com}
\affiliation{%
  \institution{Amazon Alexa AI}
  \city{Seattle}
  \state{WA}
  \country{USA}
}

\author{Eunah Cho}
\authornote{Team leaders. Authors are alphabetically ordered by last name.}
\email{eunahch@amazon.com}
\affiliation{%
  \institution{Amazon Alexa AI}
  \city{Seattle}
  \state{WA}
  \country{USA}
}

\author{Xing Fan}
\authornotemark[2]
\email{fanxing@amazon.com}
\affiliation{%
  \institution{Amazon Alexa AI}
  \city{Seattle}
  \state{WA}
  \country{USA}
}

\author{Xiaojiang Huang}
\authornotemark[2]
\email{xjhuang@amazon.com}
\affiliation{%
  \institution{Amazon Alexa AI}
  \city{Seattle}
  \state{WA}
  \country{USA}
}

\author{Yanbin Lu}
\authornotemark[2]
\email{luyanbin@amazon.com}
\affiliation{%
  \institution{Amazon Alexa AI}
  \city{Seattle}
  \state{WA}
  \country{USA}
}

\author{Aram Galstyan}
\authornote{Amazon scholar.}
\email{argalsty@amazon.com}
\affiliation{%
  \institution{Amazon Alexa AI}
  \city{Seattle}
  \state{WA}
  \country{USA}
}

\renewcommand{\shortauthors}{Zheng and Ziyan and Fan, et al.}

\begin{abstract}
Conversational AI systems like Alexa, Siri, and Google Assistant require an understanding of defective queries to ensure robust conversational functionality and minimize user friction. Such defective queries often stem from user ambiguities, errors, or inaccuracies in automatic speech recognition (ASR) and natural language understanding (NLU).

A \textit{Personalized query rewriting} (personalized QR) system strives to minimize defective queries by considering individual user behavior and preferences. It primarily depends on a user index of past successful interactions with the conversational AI. However, this method faces challenges with \textit{unseen} interactions, i.e., those novel user interactions not covered by the user's historical index. Unseen interactions can constitute half of the total user traffic volume, even when the user index is built from a year's worth of history.

This paper introduces our "\textit{Collaborative Query Rewriting}" approach, aiming at utilizing the underlying topological information to assist in rewriting defective queries arising from unseen user interactions. This approach first builds a "User \textbf{\underline{F}}eedback \textbf{\underline{I}}nteraction \textbf{\underline{G}}raph" (\textit{FIG}) of historical user-entity interactions. We then traverse the multi-hop user affinity to create an additional index, referred to as the "\textit{collaborative user index}". 

This paper then further explore the use of Large Language Models (LLMs) in conjunction with graph traversal to further boost the coverage of unseen interactions. We fine-tuned the LLM "Dolly-V2" 7B model on 200K users' affinities, leading to a significant increase in index coverage for future user interactions in comparison to only using graph traversal. This improvement, in turn, not only significantly enhances the performance of query rewriting (QR) for unseen queries, but also facilitates easier balancing between the index size and the coverage, a crucial requirement for the deployment of our approach into production. 

\end{abstract}



\maketitle

\section{Introduction}

\textit{Defective queries} frequently occur during user interactions with conversational AI systems such as Alexa, Siri or Google Assistant. These are induced by user ambiguities or mistakes, along with errors in automatic speech recognition (ASR) or natural language understanding (NLU). Defective queries impact the robustness of the conversational AI system, as they hinder users from receiving the intended results and often require further clarification. \textit{Query Rewriting} (QR) is a subsystem within the conversational AI that plays a crucial role in reducing defective queries. By automatically refining or correcting these defective queries, QR enhances the overall robustness of the AI system and significantly improves the user experience.

\textit{Personalized query rewriting} (Personalized QR) takes into account individual preferences or unique error patterns identified from a user's historical interactions with the conversational AI. It plays a crucial role in addressing a wide range of user-specific defects, particularly in the torso and tail distribution \cite{roshan2020personalized, Naresh2022PENTATRONPC, cho2021personalized, li2022query}. For instance, when a user presents a defective query like "play abcdefg", a non-personalized QR system might rewrite it to "play alphabetic song" based on the high overall transition probability from "abcdefg" to "alphabetic song". However, for this particular user, the query is intended for the song "abcdefu" by the American singer Gayle.

A \textit{search-based personalized QR} system \cite{cho2021personalized} typically requires a \textit{personalized index} to reflect historical non-defective experiences for each user (see Section \ref{sec:search_based_qr_system} and Figure \ref{fig:search_qr} for more details). We refer to a personalized index built from the user's own history as the \textit{user history index}. The user history index includes each user's own historical successful queries, rephrases, and rewrites, and related metadata \& statistics (e.g. how many times user said a query in the history). During runtime, given a user query (e.g. "play abcdefg"), the system checks if a successful historical query utterance (e.g. "play abcdefu by gale") in the user history index closely matches the current query. The degree of match is determined through methods such as elastic search retrieval or a neural similarity model. If a match is found, it is used to rewrite the initial query. The user interactions covered by the user history index are referred to as the "\textbf{\textit{seen interactions}}".

Despite the effectiveness of personalized QR for reducing defects in conversational AI, we have identified the challenge posed by \textbf{\textit{unseen interactions}} not covered by the user history index. We have observed that users frequently engage in new experiences, leading to approximately 50\% of the queries/interactions within a week-long period not being covered by the user history index, even when the index is constructed from a year-long history. We refer to these queries/interactions as "\textit{unseen}". Moreover, we observed a roughly 7\% higher defects in the unseen queries/interactions, which highlight the potential opportunity for the unseen interactions. 

\begin{table*}
\centering
\begin{tabular}{lccccc}
\hline
\textbf{n-Hop Affinity}                                       & \textbf{1}              & \textbf{2}         & \textbf{3}        & \textbf{4}         & \textbf{5}          \\
\hline
\% \textbf{Unseen Interactions Covered}             & 0\%            & 10\%      & 20\%     & 26\%      & 31\%       \\
\% \textbf{Defective Unseen Interactions Covered}   & 0\%            & 12\%      & 24\%     & 32\%      & 40\%       \\
\textbf{Avg. \# of Rewrite Candidates in The Affinity}            & \textless{}100 & $\sim$600 & $\sim$3K & $\sim$20K & $\sim$100K \\

\hline
\end{tabular}
\caption{Unseen user interaction coverage by the collaborative user index enriched by up to 5-hop user affinity in FIG. FIG is built from one-year user history and the evaluation is done on one-week interactions in the future.}
\label{tab:nhop_affiinity_coverage}
\end{table*}

We introduce our approach, "\textit{Collaborative Query Rewriting}" (Collaborative QR), which aims to address the challenge of index coverage. This approach is inspired by our observation that users who interact with similar entities through a conversational AI system often make similar queries or experience comparable defects (see Figure \ref{fig:fig_illustration}). The cornerstone of our approach is the "User \textbf{\underline{F}}eedback \textbf{\underline{I}}nteraction \textbf{\underline{G}}raph" (FIG), which captures users' previous interactions with various entities through the conversational AI in a user-entity interaction graph. Each interaction edge in the FIG corresponds to historical successful queries/rephrases made by users, successful rewrites generated by the QR system, and feedback signals from users(see Section \ref{sec:fig}). Our key idea is to leverage the FIG to form a \textit{collaborative user index} consisting of additional rewrite candidates not in the user history index.

\textit{Graph traversal} through the FIG is the most straightforward approach for constructing the collaborative user index. We search user affinity in the FIG along “user→entity→user→entity→...” paths, and employ rules to filter, rank, and select top historical non-defect query utterances in the user affinity. Table \ref{tab:nhop_affiinity_coverage} shows a few coverage statistics that support this method. We built the FIG from one-year user history, and evaluate one-week future user interactions. We find a considerable 40\% of the defective unseen interactions can be covered within the 5-hop affinity. 

While graph traversal offers notable benefits, it also introduces certain challenges. In a production setting, traversal is limited to a maximum of 3 hops due to computational limitations. Even more critical is the issue of index size. A search-based system requires careful index size management as an overly large index could not only harm the retrieval precision, but also increase the runtime system's latency. However, for the collaborative user index generated through graph traversal, it requires a fairly large index size cap of 500 for each user to achieve satisfactory coverage of unseen interactions.\footnote{The user history index only needs a much smaller size cap of 100 for each user. User history index only stores user's own historical interactions, and the cap at 100 is enough to for the index store all user historical interactions for the majority of users.}. Therefore, improving the ranking of the collaborative user index and reducing its size is an important problem to address for collaborative QR.

To enhance the collaborative user index, we look into \textit{Large Language Models} (LLMs)\cite{Brown2020LanguageMA,Touvron2023LLaMAOA,OpenAI2023GPT4TR}. LLMs have shown remarkable adaptability in various NLP tasks. However, to the best of our knowledge, there are no existing publications on leveraging LLMs to enhance the user index for query defect reduction and improve the robustness of conversational AI systems. In this paper, we explore the possibility of applying a publicly available LLM "Dolly-V2" 7B\cite{databrickslabs_dolly} in combination with graph traversal to further improve the coverage of the collaborative user index. The increased coverage would also facilitate balancing between coverage and index size cap\footnote{Note that user index can be pre-computed offline, therefore LLM inference latency is not impacting the runtime system efficiency. The approach is feasible as long as the collaborative user index can be pre-computed within reasonable time and cost, say 2 weeks with 64 A-100 GPUs}. The basic idea is straightforward - prompting the LLM with items the user has interacted with in the past, and ask the LLM to generate other items the user might be interested in the future. The method leverages the latent relational knowledge already embodied in the LLM, and breaks the limitation by the physical user affinity in the FIG graph (see the example in \ref{sec:llm_driven_rewrite_trigger}). 

Our key contributions are summarized as the following:
\begin{enumerate}[itemsep=0pt,parsep=0pt,topsep=0pt,partopsep=0pt]
    \item To the best of our knowledge, we are the first to propose "Collaborative Query Rewriting", which uses topological user-entity interaction information to reduce query defects.
    \item We show that we can achieve competitive QR performance for unseen user interactions compared to defect reduction on "seen" interactions. This is accomplished by creating a collaborative user index through graph traversal, with an index size cap for the collaborative user index at 500. We have validated our approach through production A/B testing, where it has shown significant improvements.
    \item We have explored the use of LLMs to learn from the user affinity in the FIG graph. After fine-tuning the "Dolly-V2" 7B model on 200K user affinity learning examples, we saw a significant increase in the index coverage for unseen user interactions. This led to a notable improvement in the defect reduction trigger rate, while the collaborative user index size cap is reduced from 500 to 200.
\end{enumerate}

\section{Related Works}
\label{sec:related_works}

Many previous works have been proposed for robust conversational understanding through QR. For non-personalized QR, \cite{bonadiman2019large} examined paraphrase retrieval for reducing defects in question answering tasks. \cite{yu2020few} suggested a Markov chain model for extracting users' reformulation patterns that can help identify rewrite pairs. \cite{chen2020pre} proposed a query encoder pre-trained with a large number of user rephrases and then fine-tuned for search-based QR task utilizing the query encoder. For personalized QR, \cite{roshan2020personalized} compared a retrieval model with a pointer-generator network with hierarchical attention for performing personalized rewrites within the smart home domain, \cite{fan2021search} extended search-based QR in spoken dialogue systems by combining a global layer for generic, non-personalized rewrites and a personalized layer. \cite{cho2021personalized} proposed a personalized index combining different types of user affinities, and introduced a re-ranking layer for the retrieval results.

There are a few prior works leveraging graph for query rewriting. \cite{antonellis2008simrank++} studied query rewriting for sponsored search using the historical click graph to identify rewrites that are similar to the user's query based on SimRank, where the goal is to identify more relevant ads to display to users. \cite{yuan2021graph} created a customer-query interaction graph and applied it to non-personalized QR through, first pre-training query embeddings using different graph representation learning methods Deepwalk, ConvKB, and GraphSAGE, and then fine-tuning on the QR task. 

There has been a surge of recent researches affirming LLM can learn from user affinity and make predictions or recommendations. \cite{chen2023palr} fine-tunes a LLaMA 7B model to learn from the user affinity of movie-lens dataset and the Amazon beauty dataset, and out-performs the SOTA models on the recommendation task. \cite{kang2023llms} investigates the ability of Large Language Models (LLMs) to understand user preferences and predict user ratings. The study finds that while zero-shot LLMs lag behind traditional recommender models that utilize user interaction data, they can achieve comparable or even superior performance when fine-tuned with a small fraction of the training data. \cite{cui2022m6} proposed a generative pretrained language model that serves as a unified foundation for various tasks in recommender systems, using user behavior data as plain texts and converts tasks into language understanding or generation. 

In comparison with the prior works, our "Collaborative Query Rewriting" work distinguishes itself by designing a user-entity interaction graph for personalized QR, leveraging user affinity to enrich user index. Through graph traversal, we build our production collaborative QR system that achieves a significant higher trigger, while maintaining competitive precision and runtime latency. Through LLM affinity learning, we demonstrate that collaborative QR can still further substantially enhance defect reduction.

\section{Methodology}

\subsection{User Feedback Interaction Graph}
\label{sec:fig}

\begin{figure*}[htbp]
    \centering
    \includegraphics[trim={0 0 0 3cm},width=16cm]{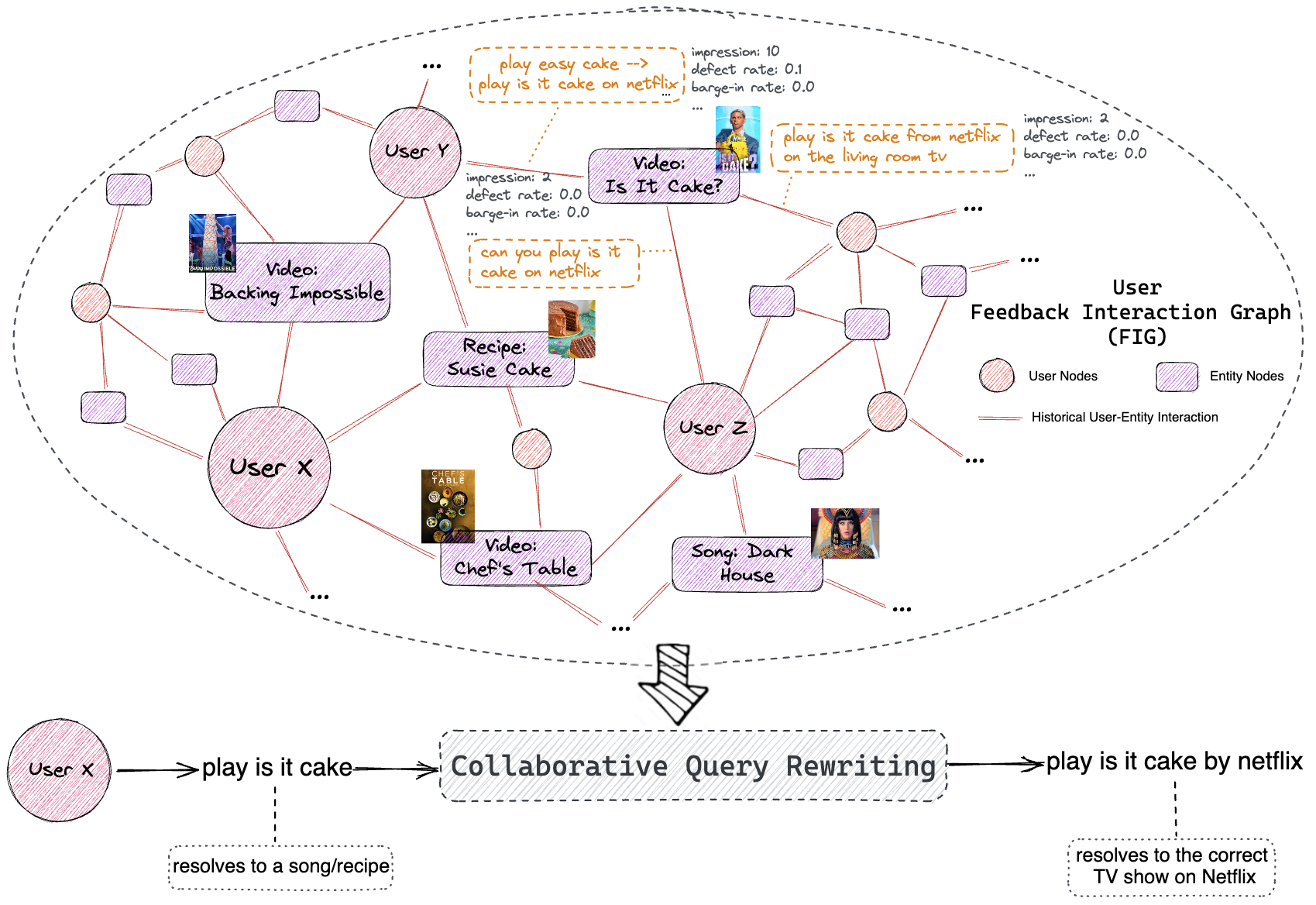}
    \caption{High-level illustration of the FIG and its application in collaborative QR. User X and Y interacted with the same or similar baking videos, which indicates their similarity. There was a successful historical rewrite "play easy cake → play is it cake on netflix" for user Y, which effectively resolved entity to the correct one. When user X encounters a defective query such as "play is it cake", the historical rewrite from user Y ("play easy cake → play is it cake on Netflix") can be considered as a rewrite candidate and utilized to correct it as "play is it cake by Netflix".}
    \label{fig:fig_illustration}
\end{figure*}

A conversational AI system requires user profiles and historical behavioral data for its personalization features. \textit{Graph} emerges as a natural structure to represent user historical interactions with various entities through the conversational AI. These entities can span diverse categories like songs, videos, books, and more. We extract non-defective user-entity interactions from the raw conversational AI logs and integrate them into a user-entity interaction graph. Different nodes of the graph represent different users and entities, while the edges encapsulate the information related to their interactions. This \textit{interaction information} encompasses the user's queries as well as associated feedback signals (e.g. impression, defect rate, barge-in rate, termination-rate). We refer to this graph as "\textit{User Feedback Interaction Graph}" (FIG), where the term “feedback” emphasizes that the graph incorporates explicit and implicit feedback from users. 

Figure \ref{fig:fig_illustration} offers a high-level depiction of the FIG and its application in collaborative QR. It includes user nodes (such as "User X", "User Y", "User Z") and entity nodes (like "Video: Is It Cake" and "Recipe: Susie Cake"). The user queries encapsulated in the edges represent \textit{non-defective} interactions between the user and the entity. Here, "non-defective" refers to user-entity interactions where the defect rate\cite{gupta2021robertaiq} falls below a certain threshold. These queries might consist of the user's original input utterances (for example, "play is it cake from netflix on the living room tv"\footnote{All queries are in lower case for this paper.}) or a pair of utterances if a rewrite for the original input was successful in the past (such as "play easy cake" being revised to "play is it cake on netflix"). Feedback signals, also encapsulated in the edges, include various elements such as impression (representing the past frequency of the query), defect rate\cite{gupta2021robertaiq}, barge-in rate (the probability that the user interrupted the agent's response to this query), termination-rate (the probability that the user stopped the agent's response to this query).


\subsection{Collaborative User Index Through Graph Traversal}
\label{sec:affinity_based_personalized_index_expansion}

We leverage the FIG to build a collaborative user index through graph traversal. The intuition is that users who have interacted with the same entities in the past are likely similar, and could also exhibit similar interactions in the future. As we traverse the FIG, we collect the interaction information encapsulated within these edges (e.g. historical queries for this interaction, with their associated feedback signals, see Section \ref{sec:fig}) and integrate them into our collaborative user index. Within this process, user queries are considered potential candidates for rewriting, while feedback signals can serve as ranking features (see Section \ref{sec:search_based_qr_system}). Currently, we limit our consideration to a 3-hop traversal (that is, paths such as "User X → Entity A → User Y → Entity B") due to computational resource constraints. The collaborative user index is pre-constructed offline to reduce runtime latency and is periodically refreshed. We also implement heuristic rules to surface more promising candidates while controlling the size of the collaborative user index (see Appendix \ref{sec:appendix_x}). 

After jointly considering runtime system constraints and unseen interaction coverage, we currently choose 500 as the collaborative user index size cap. 

\subsection{Collaborative User Index Enhanced By LLM}
\label{sec:user_index_by_llm}

Large language models have showcased remarkable capability in deducing user preferences and predicting future behavior by analyzing historical interactions\cite{chen2023palr}. Before the graph traversal step, we employ a large language model for link prediction between the user nodes and the entity nodes. We have chosen the "Dolly-V2" model and apply instruction-based fine-tuning, a proven effective method in recent developments of large language models (LLMs)\cite{alpaca, Wei2021FinetunedLM}. Currently, our exploration is focused on the Music/Video domains (they are dominant domains taking about 80\% of total user traffic volume), where we aim to leverage the factual knowledge stored within the LLM's parameters.

To perform fine-tuning, we utilize the user's historical interacted entities as training input, which corresponds to the user's 1-hop connected nodes in the FIG. 
The training labels for the model consist of the entities that the same user interacted with during the subsequent month following the training input.
Here are the examples of the training data:

\begin{small}
\begin{center} 
    \fbox{%
        \parbox{0.9\columnwidth}{
            \textbf{Instruction}: Recommend 10 other movies based on the user's watching history.
            
            \textbf{Input}: The user watched movies "Pink Floyd - The Wall", "Canadian Bacon", "G.I. Jane", "Across the Universe", ..., "Down by Law".

            \textbf{Label}: "Almost Famous", "Full Metal Jacket", "The Hurt Locker", ...
        }
    }
\end{center}
\begin{center} 
    \fbox{%
        \parbox{0.9\columnwidth}{
            \textbf{Instruction}: Recommend ten other songs based on the user's listening history.

            \textbf{Input}: The user listened to songs "Jolene by Dolly Patron", "I Walk the Line by Johnny Cash", "Ring of Fire by Johnny Cash", ..., "Take Me Home, Country Roads by John Denver".

            \textbf{Label}: "Fancy by Reba McEntire", "Sweet Dreams by Patsy Cline", "Coat of Many Colors", ...
        }
    }
\end{center}
\end{small}

\vspace{0.2cm}

At the inference stage, the LLM can infer potential edges between a user node and entity nodes that are not currently connected to the user node in the FIG. Then these predicted potential edges are utilized in the graph traversal through paths like "User X → Predicted Entity A → User Y". We collect rewrite candidates and their associated features on the "Predicted Entity A → User Y" edges to enrich the User X's collaborative index.

\subsection{Search-Based Collaborative QR System}
\label{sec:search_based_qr_system}

\begin{figure}[H]
    \centering
    \includegraphics[trim={0 0 0 3cm},width=\columnwidth]{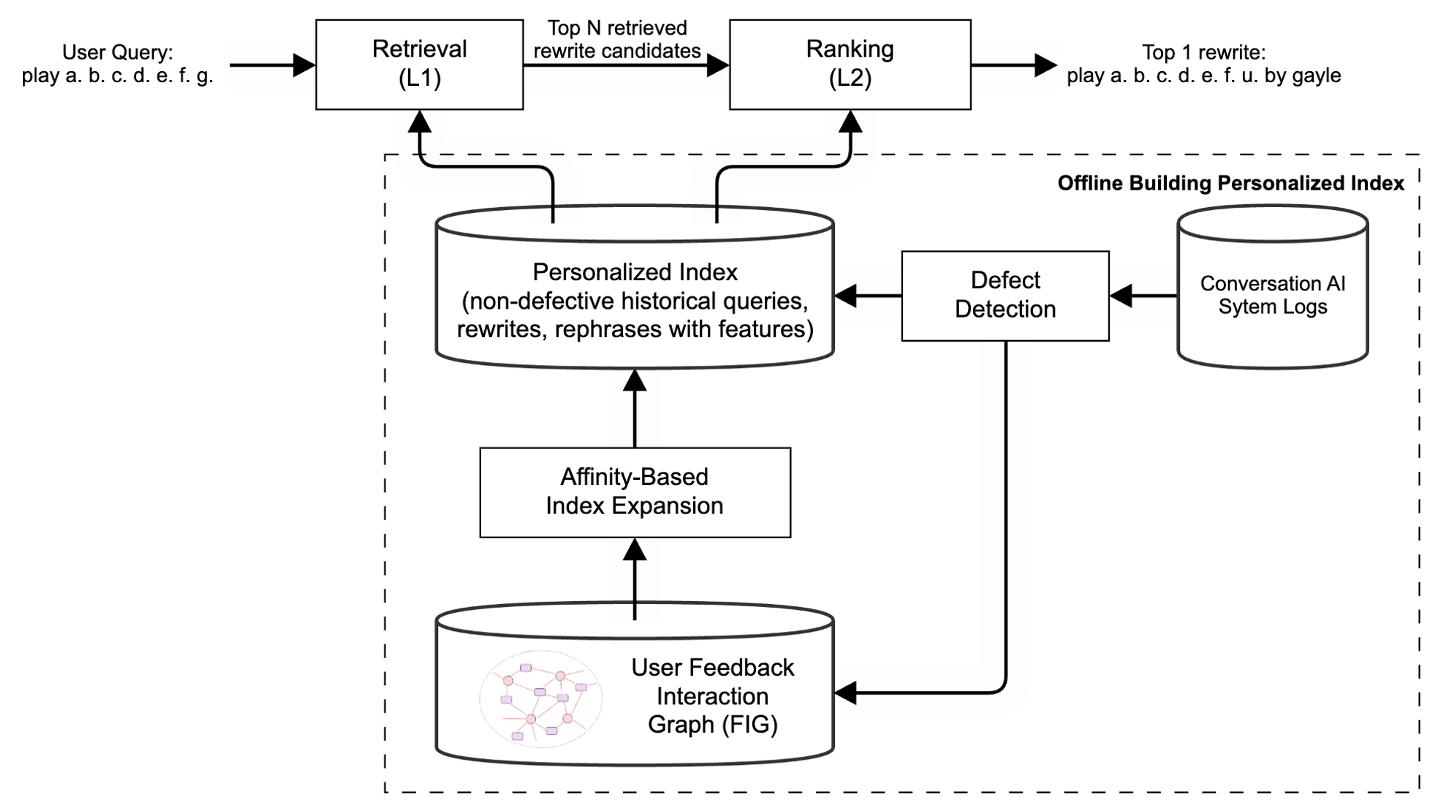}
    \caption{The high-level workflow of our search-based collaborative query rewriting system.}
    \label{fig:search_qr}
\end{figure}

Our collaborative search-based QR system, similar to previous search-based QR systems\cite{fan2021search,cho2021personalized,cai2023kg}, follows a two-stage design consisting of a retrieval module (L1) and ranking module (L2), as illustrated in Figure \ref{fig:search_qr}. The \textit{personalized index} serves as both the search space and ranking feature store. In our system, the personalized index includes the user history index plus the collaborative user index, as described by Section \ref{sec:affinity_based_personalized_index_expansion} and Section \ref{sec:user_index_by_llm}.

\vspace{0.7cm}

\vspace{-0.4cm}

The retrieval module in our collaborative QR system aims to retrieve a set of relevant rewrite candidates from the personalized index. The goal is to maximize recall with low latency and computational cost. Our production system uses a Transformer-based model as the utterance encoder, by taking a similar approach as \cite{chen2020pre}\footnote{We stack Transformer layers as the L1 encoder model that can run within the latency budget. The runtime system has strict latency requirement and cannot yet host a larger model like BERT.}. The learning objective of the retrieval module is to project the embedding of input query and that of target rewrite closely.

After retrieving potential rewrite candidates, the ranking module leverages a gradient boosting ranker model to select the most suitable rewrite. The current ranker incorporates various aforementioned feedback signals as features. These features are calculated at the global level and the user level. For example, the user level impression feature counts the number of times the query appears in the user's history, while the global level impression feature indicates its occurrence across all users' histories.

The collaborative user index expands search space. While this expansion introduces more opportunities, it also introduces more noise. Table \ref{tab:ranking_result} row \#2 shows the QR quality is significantly harmed by the increased search space. To resolve this issue, we adopt a strategy of increasing the size of the encoder by stacking more Transformer layers in the retrieval module. We further incorporate \textit{guardrail features} and \textit{graph-based features} to address false triggers and ensure system precision (see Appendix \ref{sec:appendix_y}).

\begin{table*}[ht]
\resizebox{\textwidth}{!}{
\begin{tabular}{c|l|ccccc}
\hline
\multirow{2}{*}{\textbf{\#}} & \multirow{2}{*}{\textbf{Ranking}} & \multicolumn{2}{c}{\textbf{\begin{tabular}[c]{@{}c@{}}Opportunity Test Set\\ {\small (Seen Interactions)}\end{tabular}}} & \multicolumn{2}{c}{\textbf{\begin{tabular}[c]{@{}c@{}}Opportunity Test Set \\  {\small (Unseen Interactions)} \end{tabular}}} & \textbf{Guardrail Test Set} \\
 & & \textbf{p@1}                                                       & \textbf{Trigger Rt.}                                                   & \textbf{p@1}                                                       & \textbf{Trigger Rt.}                                                     & \textbf{False Trigger Rt.}          \\
\hline
1 & \textbf{Personalized QR(Baseline)} & 82.0\%  & 79.5\%  & N/A   & N/A  & 10.4\%                      \\
\hline
2 & \textbf{Collaborative QR}                                                                          & 78.3\%                                                    & \textbf{82.4}\%                                                    & 74.5\%                                                     & 4.77\%                                                      & 12.5\%                      \\
3 & \textbf{+ L1 Encoder More Transformer layers}                                                                          & 80.2\%                                                    & 80.9\%                                                    & 76.5\%                                                     & 4.82\%                                                      & 8.6\%                       \\
4 & \textbf{+ L2 guardrail/graph-based features}                                                                     & \textbf{85.2}\%                                                    & 81.5\%                                                    & \textbf{83.1}\%                                                     & \textbf{5.01}\%                                                      & \textbf{2.1}\%                      \\
\hline
\end{tabular}}
\caption{Collaborative QR evaluation results (L1 retrieval + L2 ranking) on the test sets. The results verify our essential idea of Collaborative QR, and affirm competitive rewrite quality can be achieved for the unseen interactions, even when the collaborative user index size cap is much larger.}
\label{tab:ranking_result}
\end{table*}

\section{Experiments}

In this section, we first demonstrate the validity our essential idea of collaborative QR, that the collaborative user index built through graph traversal can already boost defect reduction, and we are able to achieve competitive precision performance with the much enlarged index after applying techniques mentioned in Section \ref{sec:search_based_qr_system}. 

After that, we demonstrate the potential of applying LLM to significantly further boost collaborative user index coverage even with a smaller index size cap, and thereby introduce 5 to 6 times more defect reduction.

\subsection{Collaborative QR With Graph Traversal}
\label{sec:exp_graph_traversal}

\subsubsection{Data}
\label{sec:exp_graph_traversal_data}
The offline evaluation of our graph-traversal based collaborative QR system includes two \textit{opportunity test sets} and one \textit{guardrail test set}. 
\begin{itemize}
    \item The opportunity test sets are weak-labeled data similar to previous works\cite{fan2021search,cho2021personalized}. We begin by identifying pairs of consecutive user utterances, where the first turn is defective but the second turn is successful. We utilize a defect detection model\cite{gupta2021robertaiq} to determine whether an utterance is defective or not. To minimize potential noise in the data and identify pairs where the second utterance is indeed a rephrase of the first utterance, we apply additional filters such as edit-distance and ASR n-best filters. Finally, the second utterance in the pair will be used as the rewrite label for the first utterance.
    We create two opportunity test sets: 1) \textit{Seen Interactions}: the rewrite label exists in the user's own history; 2) \textit{Unseen Interactions}: the rewrite label is not found in the user's history.
    \item The guardrail test set consists of historically successful user query utterances. 
    The QR system should not trigger rewrite for any test case in the guardrail test set.  

\end{itemize}

\subsubsection{Evaluation metrics}
For opportunity test sets, we use precision and trigger rate as metrics. Precision measures how often the triggered top 1 rewrite's NLU hypothesis matches the rewrite label's NLU hypothesis. Trigger rate represents the ratio between rewrite-triggered test cases and all test cases. The QR component is triggered when the prediction score of the top 1 rewrite is above an empirically chosen threshold.

For the guardrail test set, we utilize the false trigger rate as the metric. This rate also represents the ratio between the number of test cases that trigger a rewrite and the total number of test cases. However, in the guardrail test set, these cases should not be triggered. Therefore a lower false trigger rate is indicative of a better query rewriting system.

\subsubsection{Offline Evaluation Results}
Table \ref{tab:ranking_result} shows the performance of collaborative QR on the test sets. As indicated by \#1 and \#2, collaborative QR is capable of enabling rewrites on the "unseen interactions" test set. However, the precision performance drops significantly (74.5\% compared to 82.0\%) due to the much larger search space of the collaborative user index, leading to a higher rate of false triggers. To mitigate this performance degradation, as discussed in Section \ref{sec:search_based_qr_system}, we introduce a larger utterance encoder to the L1 retrieval and incorporate guardrail features to the L2 ranking. Following these improvements, as shown by \#3 and \#4, we achieve competitive precision performance (83.1\% compared to 82.0\%). Furthermore, we notice a substantial reduction in the false trigger rate on the guardrail test set (10.4\% reduced to 2.1\%).

\vspace{-0.1cm}

\subsubsection{Online Evaluation Results}
We deployed our collaborative QR system and evaluated its online performance. Overall it significantly introduces \textbf{23\%} additional personalized defect removal. In the A/B experiment, we observed significant \textbf{19\%} relative reduction of defect rate with p-value<0.0001. 

\subsection{Collaborative QR Enhanced By LLMs}

\subsubsection{Collaborative user index coverage} Table \ref{tab:user_index_coverage} shows the coverage of unseen interactions by collaborative user indexes constructed using different methods. We evaluate the coverage on two dominant domains Video and Music ($\sim$80\% of traffic volume). Notably, the fine-tuned Dolly V2 enhanced collaborative user index significant exceeds the performance of the graph traversal only method. The Dolly V2 enhanced index with index size 200 significantly outperforms graph-traversal based index with size cap 500.

\begin{table}[]
\resizebox{\columnwidth}{!}{%
\begin{tabular}{r|ccc|ccc}
\multicolumn{1}{l}{}   & \multicolumn{3}{c}{\textbf{Video}}         & \multicolumn{3}{c}{\textbf{Music}}         \\
\textbf{Index Construction Method}              & \textbf{100} & \textbf{200} & \textbf{500} & \textbf{100} & \textbf{200} & \textbf{500} \\
\hline
Graph Traversal Only        & 1.8\%        & 3.8\%        & 6.3\%        & 1.1\%        & 2.7\%        & 5.4\%          \\
\hline
\begin{tabular}[c]{@{}r@{}}+Dolly-V2 Link Prediction\\ (not fine-tuned)\end{tabular} & 2.5\%        & 5.3\%        & N/A          & 1.4\%        & 3.6\%        & N/A          \\
\hline
\begin{tabular}[c]{@{}r@{}}+Dolly-V2 Link Prediction\\ (fine-tuned)\end{tabular}    & \textbf{10.8}\%       & \textbf{24.5}\%       & N/A          & \textbf{8.5}\%        & \textbf{18.4}\%       & N/A     
\\
\hline
\end{tabular}%
}
\caption{Comparison of \underline{unseen user interaction coverage} by collaborative user indexes constructed by different methods, with \underline{index size cap being 100, 200 and 500}. }
\label{tab:user_index_coverage}
\end{table}

\subsubsection{Offline Evaluation Results} Table \ref{tab:llm_qr} shows QR performance for using collaborative user indexes constructed by different methods, with collaborative user index size limited to 200. LLM-enhanced collaborative user index achieves significantly higher trigger rate due to its much higher coverage. 

\subsubsection{An example of LLM-driven rewrite trigger.} 
\label{sec:llm_driven_rewrite_trigger}
A user likes playing musicals, and has historically listened to musicals such as "My Fair Lady", "The Sound of Music", "Hamilton", etc. The user interacted another musical "Guys and Dolls" in the next week. This musical was not in the 3-hop user affinity in FIG (but in the 5-hop affinity) and hence not captured by the graph traversal. The LLM infers this potential future affinity. As a result, we covered a defective query with ASR error "could you gods dolls" and rewrite it as "play guys and dolls"

To get some insight, we particularly asked the fine-tuned Dolly-V2 7B model the following question.

\begin{center} 
    \fbox{%
        \parbox{0.8\columnwidth}{
            User previously listened to "My Fair Lady", "The Sound of Music", "Hamilton", ... Why would you recommend "Guys and Dolls"? 
        }
    }
\end{center}

It responded

\begin{center} 
    \fbox{%
        \parbox{0.8\columnwidth}{
            "Guys and Dolls" is a classic well-known musical in the same genre.
        }
    }
\end{center}

This example serves to illustrate that there exists an inherent knowledge graph within the parameters of the LLM. By observing user affinity, the LLM could utilize this internal knowledge to infer user preferences that may extend beyond the physical topology of the user-entity interaction graph.

\begin{table}[]
\resizebox{\columnwidth}{!}{%
\begin{tabular}{r|ccc|ccc}
\multicolumn{1}{l}{}                                                                           & \multicolumn{3}{c}{\textbf{Video}}                                                                  & \multicolumn{3}{c}{\textbf{Music}}                                                                                      \\
\textbf{Index Construction Method}                                                                         & \textbf{p@1} & \textbf{trigger} & \textbf{\begin{tabular}[c]{@{}c@{}}false \\ trigger\end{tabular}} & \multicolumn{1}{r}{\textbf{p@1}} & \textbf{trigger} & \textbf{\begin{tabular}[c]{@{}c@{}}false \\ trigger\end{tabular}} \\
\hline
Graph Traversal Only                                                                                            & \textbf{81.5}\%         & 3.7\%           & 2.7\%                                                             & 79.7\%                           & 2.2\%           & 1.8\%                                                             \\
\hline
\begin{tabular}[c]{@{}r@{}}+Dolly-V2   Link Prediction\\      (fine-tuned)\end{tabular}                   & 81.3\%         & \textbf{19.6}\%           & \textbf{2.2}\%                                                             & \textbf{81.5}\%                           & \textbf{15.4}\%           & \textbf{1.7}\%                                                             \\

\hline
\end{tabular}%
}
\caption{Comparison of the QR performance for unseen user interactions, using collaborative user indexes built using different methods. Collaborative index size is limited to 200. }
\label{tab:llm_qr}
\end{table}

\section{Conclusion}

In this paper, we first identify a large volume of unseen user interactions occur every week with higher rate of defective queries. We then propose the "Collaborative Query Rewriting" approach that aims to reduce these defects specifically in unseen interactions. Performance degradation due to an enlarged index was rectified by implementing additional transformer layers for the L1 retrieval model and incorporating guardrail and graph features in the L2 ranking model.

Furthermore, we investigated the potential of an LLM in enhancing the collaborative QR approach. We found great potential for an LLM to significantly improve the coverage of the collaborative user index that can lead to a significant 5 to 6 times more reduction of query defects. As a future course of action, we aim to experiment techniques such as distillation, teacher models, etc. to further optimize performance.

\begin{acks}

\end{acks}

\bibliographystyle{ACM-Reference-Format}
\bibliography{custom}


\begin{thebibliography}{21}


\ifx \showCODEN    \undefined \def \showCODEN     #1{\unskip}     \fi
\ifx \showDOI      \undefined \def \showDOI       #1{#1}\fi
\ifx \showISBNx    \undefined \def \showISBNx     #1{\unskip}     \fi
\ifx \showISBNxiii \undefined \def \showISBNxiii  #1{\unskip}     \fi
\ifx \showISSN     \undefined \def \showISSN      #1{\unskip}     \fi
\ifx \showLCCN     \undefined \def \showLCCN      #1{\unskip}     \fi
\ifx \shownote     \undefined \def \shownote      #1{#1}          \fi
\ifx \showarticletitle \undefined \def \showarticletitle #1{#1}   \fi
\ifx \showURL      \undefined \def \showURL       {\relax}        \fi
\providecommand\bibfield[2]{#2}
\providecommand\bibinfo[2]{#2}
\providecommand\natexlab[1]{#1}
\providecommand\showeprint[2][]{arXiv:#2}

\bibitem[Antonellis et~al\mbox{.}(2008)]%
        {antonellis2008simrank++}
\bibfield{author}{\bibinfo{person}{Ioannis Antonellis}, \bibinfo{person}{Hector
  Garcia-Molina}, {and} \bibinfo{person}{Chi-Chao Chang}.}
  \bibinfo{year}{2008}\natexlab{}.
\newblock \showarticletitle{Simrank++ query rewriting through link analysis of
  the clickgraph (poster)}. In \bibinfo{booktitle}{\emph{Proceedings of the
  17th international conference on World Wide Web}}.
  \bibinfo{pages}{1177--1178}.
\newblock


\bibitem[Bonadiman et~al\mbox{.}(2019)]%
        {bonadiman2019large}
\bibfield{author}{\bibinfo{person}{Daniele Bonadiman},
  \bibinfo{person}{Anjishnu Kumar}, {and} \bibinfo{person}{Arpit Mittal}.}
  \bibinfo{year}{2019}\natexlab{}.
\newblock \showarticletitle{Large scale question paraphrase retrieval with
  smoothed deep metric learning}.
\newblock \bibinfo{journal}{\emph{arXiv preprint arXiv:1905.12786}}
  (\bibinfo{year}{2019}).
\newblock


\bibitem[Brown et~al\mbox{.}(2020)]%
        {Brown2020LanguageMA}
\bibfield{author}{\bibinfo{person}{Tom~B. Brown}, \bibinfo{person}{Benjamin
  Mann}, \bibinfo{person}{Nick Ryder}, \bibinfo{person}{Melanie Subbiah},
  \bibinfo{person}{Jared Kaplan}, \bibinfo{person}{Prafulla Dhariwal},
  \bibinfo{person}{Arvind Neelakantan}, \bibinfo{person}{Pranav Shyam},
  \bibinfo{person}{Girish Sastry}, \bibinfo{person}{Amanda Askell},
  \bibinfo{person}{Sandhini Agarwal}, \bibinfo{person}{Ariel Herbert-Voss},
  \bibinfo{person}{Gretchen Krueger}, \bibinfo{person}{T.~J. Henighan},
  \bibinfo{person}{Rewon Child}, \bibinfo{person}{Aditya Ramesh},
  \bibinfo{person}{Daniel~M. Ziegler}, \bibinfo{person}{Jeff Wu},
  \bibinfo{person}{Clemens Winter}, \bibinfo{person}{Christopher Hesse},
  \bibinfo{person}{Mark Chen}, \bibinfo{person}{Eric Sigler},
  \bibinfo{person}{Mateusz Litwin}, \bibinfo{person}{Scott Gray},
  \bibinfo{person}{Benjamin Chess}, \bibinfo{person}{Jack Clark},
  \bibinfo{person}{Christopher Berner}, \bibinfo{person}{Sam McCandlish},
  \bibinfo{person}{Alec Radford}, \bibinfo{person}{Ilya Sutskever}, {and}
  \bibinfo{person}{Dario Amodei}.} \bibinfo{year}{2020}\natexlab{}.
\newblock \showarticletitle{Language Models are Few-Shot Learners}.
\newblock \bibinfo{journal}{\emph{ArXiv}}  \bibinfo{volume}{abs/2005.14165}
  (\bibinfo{year}{2020}).
\newblock


\bibitem[Cai et~al\mbox{.}(2023)]%
        {cai2023kg}
\bibfield{author}{\bibinfo{person}{Jinglun Cai}, \bibinfo{person}{Mingda Li},
  \bibinfo{person}{Ziyan Jiang}, \bibinfo{person}{Eunah Cho},
  \bibinfo{person}{Zheng Chen}, \bibinfo{person}{Yang Liu},
  \bibinfo{person}{Xing Fan}, {and} \bibinfo{person}{Chenlei Guo}.}
  \bibinfo{year}{2023}\natexlab{}.
\newblock \showarticletitle{KG-ECO: Knowledge Graph Enhanced Entity Correction
  For Query Rewriting}. In \bibinfo{booktitle}{\emph{ICASSP 2023-2023 IEEE
  International Conference on Acoustics, Speech and Signal Processing
  (ICASSP)}}. IEEE, \bibinfo{pages}{1--5}.
\newblock


\bibitem[Chen(2023)]%
        {chen2023palr}
\bibfield{author}{\bibinfo{person}{Zheng Chen}.}
  \bibinfo{year}{2023}\natexlab{}.
\newblock \showarticletitle{PALR: Personalization Aware LLMs for
  Recommendation}.
\newblock \bibinfo{journal}{\emph{arXiv preprint arXiv:2305.07622}}
  (\bibinfo{year}{2023}).
\newblock


\bibitem[Chen et~al\mbox{.}(2020)]%
        {chen2020pre}
\bibfield{author}{\bibinfo{person}{Zheng Chen}, \bibinfo{person}{Xing Fan},
  {and} \bibinfo{person}{Yuan Ling}.} \bibinfo{year}{2020}\natexlab{}.
\newblock \showarticletitle{Pre-training for query rewriting in a spoken
  language understanding system}. In \bibinfo{booktitle}{\emph{ICASSP 2020-2020
  IEEE International Conference on Acoustics, Speech and Signal Processing
  (ICASSP)}}. IEEE, \bibinfo{pages}{7969--7973}.
\newblock


\bibitem[Cho et~al\mbox{.}(2021)]%
        {cho2021personalized}
\bibfield{author}{\bibinfo{person}{Eunah Cho}, \bibinfo{person}{Ziyan Jiang},
  \bibinfo{person}{Jie Hao}, \bibinfo{person}{Zheng Chen},
  \bibinfo{person}{Saurabh Gupta}, \bibinfo{person}{Xing Fan}, {and}
  \bibinfo{person}{Chenlei Guo}.} \bibinfo{year}{2021}\natexlab{}.
\newblock \showarticletitle{Personalized search-based query rewrite system for
  conversational ai}. In \bibinfo{booktitle}{\emph{Proceedings of the 3rd
  Workshop on Natural Language Processing for Conversational AI}}.
  \bibinfo{pages}{179--188}.
\newblock


\bibitem[Cui et~al\mbox{.}(2022)]%
        {cui2022m6}
\bibfield{author}{\bibinfo{person}{Zeyu Cui}, \bibinfo{person}{Jianxin Ma},
  \bibinfo{person}{Chang Zhou}, \bibinfo{person}{Jingren Zhou}, {and}
  \bibinfo{person}{Hongxia Yang}.} \bibinfo{year}{2022}\natexlab{}.
\newblock \showarticletitle{M6-Rec: Generative Pretrained Language Models are
  Open-Ended Recommender Systems}.
\newblock \bibinfo{journal}{\emph{arXiv preprint arXiv:2205.08084}}
  (\bibinfo{year}{2022}).
\newblock


\bibitem[Fan et~al\mbox{.}(2021)]%
        {fan2021search}
\bibfield{author}{\bibinfo{person}{Xing Fan}, \bibinfo{person}{Eunah Cho},
  \bibinfo{person}{Xiaojiang Huang}, {and} \bibinfo{person}{Edward Guo}.}
  \bibinfo{year}{2021}\natexlab{}.
\newblock \showarticletitle{Search based self-learning query rewrite system in
  conversational ai}.
\newblock  (\bibinfo{year}{2021}).
\newblock


\bibitem[Gupta et~al\mbox{.}(2021)]%
        {gupta2021robertaiq}
\bibfield{author}{\bibinfo{person}{Saurabh Gupta}, \bibinfo{person}{Xing Fan},
  \bibinfo{person}{Derek Liu}, \bibinfo{person}{Benjamin Yao},
  \bibinfo{person}{Yuan Ling}, \bibinfo{person}{Kun Zhou},
  \bibinfo{person}{Tuan-Hung Pham}, {and} \bibinfo{person}{Edward Guo}.}
  \bibinfo{year}{2021}\natexlab{}.
\newblock \showarticletitle{Robertaiq: An efficient framework for automatic
  interaction quality estimation of dialogue systems}.
\newblock  (\bibinfo{year}{2021}).
\newblock


\bibitem[Kang et~al\mbox{.}(2023)]%
        {kang2023llms}
\bibfield{author}{\bibinfo{person}{Wang-Cheng Kang}, \bibinfo{person}{Jianmo
  Ni}, \bibinfo{person}{Nikhil Mehta}, \bibinfo{person}{Maheswaran
  Sathiamoorthy}, \bibinfo{person}{Lichan Hong}, \bibinfo{person}{Ed Chi},
  {and} \bibinfo{person}{Derek~Zhiyuan Cheng}.}
  \bibinfo{year}{2023}\natexlab{}.
\newblock \showarticletitle{Do LLMs Understand User Preferences? Evaluating
  LLMs On User Rating Prediction}.
\newblock \bibinfo{journal}{\emph{arXiv preprint arXiv:2305.06474}}
  (\bibinfo{year}{2023}).
\newblock


\bibitem[Labs(2021)]%
        {databrickslabs_dolly}
\bibfield{author}{\bibinfo{person}{Databricks Labs}.}
  \bibinfo{year}{2021}\natexlab{}.
\newblock \bibinfo{title}{Dolly: A Tool for Data Generation and Benchmarking}.
\newblock
\newblock
\urldef\tempurl%
\url{https://github.com/databrickslabs/dolly}
\showURL{%
\tempurl}
\newblock
\shownote{Available at: \url{https://github.com/databrickslabs/dolly}}.


\bibitem[Li et~al\mbox{.}(2022)]%
        {li2022query}
\bibfield{author}{\bibinfo{person}{Sen Li}, \bibinfo{person}{Fuyu Lv},
  \bibinfo{person}{Taiwei Jin}, \bibinfo{person}{Guiyang Li},
  \bibinfo{person}{Yukun Zheng}, \bibinfo{person}{Tao Zhuang},
  \bibinfo{person}{Qingwen Liu}, \bibinfo{person}{Xiaoyi Zeng},
  \bibinfo{person}{James Kwok}, {and} \bibinfo{person}{Qianli Ma}.}
  \bibinfo{year}{2022}\natexlab{}.
\newblock \showarticletitle{Query Rewriting in TaoBao Search}. In
  \bibinfo{booktitle}{\emph{Proceedings of the 31st ACM International
  Conference on Information \& Knowledge Management}}.
  \bibinfo{pages}{3262--3271}.
\newblock


\bibitem[Naresh et~al\mbox{.}(2022)]%
        {Naresh2022PENTATRONPC}
\bibfield{author}{\bibinfo{person}{Niranjan~Uma Naresh}, \bibinfo{person}{Ziyan
  Jiang}, \bibinfo{person}{Ankit}, \bibinfo{person}{Sungjin Lee},
  \bibinfo{person}{Jie Hao}, \bibinfo{person}{Xing Fan}, {and}
  \bibinfo{person}{Chenlei Guo}.} \bibinfo{year}{2022}\natexlab{}.
\newblock \showarticletitle{PENTATRON: PErsonalized coNText-Aware Transformer
  for Retrieval-based cOnversational uNderstanding}. In
  \bibinfo{booktitle}{\emph{Conference on Empirical Methods in Natural Language
  Processing}}.
\newblock


\bibitem[OpenAI(2023)]%
        {OpenAI2023GPT4TR}
\bibfield{author}{\bibinfo{person}{OpenAI}.} \bibinfo{year}{2023}\natexlab{}.
\newblock \showarticletitle{GPT-4 Technical Report}.
\newblock \bibinfo{journal}{\emph{ArXiv}}  \bibinfo{volume}{abs/2303.08774}
  (\bibinfo{year}{2023}).
\newblock


\bibitem[Roshan-Ghias et~al\mbox{.}(2020)]%
        {roshan2020personalized}
\bibfield{author}{\bibinfo{person}{Alireza Roshan-Ghias},
  \bibinfo{person}{Clint~Solomon Mathialagan}, \bibinfo{person}{Pragaash
  Ponnusamy}, \bibinfo{person}{Lambert Mathias}, {and} \bibinfo{person}{Chenlei
  Guo}.} \bibinfo{year}{2020}\natexlab{}.
\newblock \showarticletitle{Personalized query rewriting in conversational ai
  agents}.
\newblock \bibinfo{journal}{\emph{arXiv preprint arXiv:2011.04748}}
  (\bibinfo{year}{2020}).
\newblock


\bibitem[Taori et~al\mbox{.}(2023)]%
        {alpaca}
\bibfield{author}{\bibinfo{person}{Rohan Taori}, \bibinfo{person}{Ishaan
  Gulrajani}, \bibinfo{person}{Tianyi Zhang}, \bibinfo{person}{Yann Dubois},
  \bibinfo{person}{Xuechen Li}, \bibinfo{person}{Carlos Guestrin},
  \bibinfo{person}{Percy Liang}, {and} \bibinfo{person}{Tatsunori~B.
  Hashimoto}.} \bibinfo{year}{2023}\natexlab{}.
\newblock \bibinfo{title}{Stanford Alpaca: An Instruction-following LLaMA
  model}.
\newblock
  \bibinfo{howpublished}{\url{https://github.com/tatsu-lab/stanford_alpaca}}.
\newblock


\bibitem[Touvron et~al\mbox{.}(2023)]%
        {Touvron2023LLaMAOA}
\bibfield{author}{\bibinfo{person}{Hugo Touvron}, \bibinfo{person}{Thibaut
  Lavril}, \bibinfo{person}{Gautier Izacard}, \bibinfo{person}{Xavier
  Martinet}, \bibinfo{person}{Marie-Anne Lachaux},
  \bibinfo{person}{Timoth{\'e}e Lacroix}, \bibinfo{person}{Baptiste
  Rozi{\`e}re}, \bibinfo{person}{Naman Goyal}, \bibinfo{person}{Eric Hambro},
  \bibinfo{person}{Faisal Azhar}, \bibinfo{person}{Aur'elien Rodriguez},
  \bibinfo{person}{Armand Joulin}, \bibinfo{person}{Edouard Grave}, {and}
  \bibinfo{person}{Guillaume Lample}.} \bibinfo{year}{2023}\natexlab{}.
\newblock \showarticletitle{LLaMA: Open and Efficient Foundation Language
  Models}.
\newblock \bibinfo{journal}{\emph{ArXiv}}  \bibinfo{volume}{abs/2302.13971}
  (\bibinfo{year}{2023}).
\newblock


\bibitem[Wei et~al\mbox{.}(2021)]%
        {Wei2021FinetunedLM}
\bibfield{author}{\bibinfo{person}{Jason Wei}, \bibinfo{person}{Maarten Bosma},
  \bibinfo{person}{Vincent Zhao}, \bibinfo{person}{Kelvin Guu},
  \bibinfo{person}{Adams~Wei Yu}, \bibinfo{person}{Brian Lester},
  \bibinfo{person}{Nan Du}, \bibinfo{person}{Andrew~M. Dai}, {and}
  \bibinfo{person}{Quoc~V. Le}.} \bibinfo{year}{2021}\natexlab{}.
\newblock \showarticletitle{Finetuned Language Models Are Zero-Shot Learners}.
\newblock \bibinfo{journal}{\emph{ArXiv}}  \bibinfo{volume}{abs/2109.01652}
  (\bibinfo{year}{2021}).
\newblock


\bibitem[Yu et~al\mbox{.}(2020)]%
        {yu2020few}
\bibfield{author}{\bibinfo{person}{Shi Yu}, \bibinfo{person}{Jiahua Liu},
  \bibinfo{person}{Jingqin Yang}, \bibinfo{person}{Chenyan Xiong},
  \bibinfo{person}{Paul Bennett}, \bibinfo{person}{Jianfeng Gao}, {and}
  \bibinfo{person}{Zhiyuan Liu}.} \bibinfo{year}{2020}\natexlab{}.
\newblock \showarticletitle{Few-shot generative conversational query
  rewriting}. In \bibinfo{booktitle}{\emph{Proceedings of the 43rd
  International ACM SIGIR conference on research and development in Information
  Retrieval}}. \bibinfo{pages}{1933--1936}.
\newblock


\bibitem[Yuan et~al\mbox{.}(2021)]%
        {yuan2021graph}
\bibfield{author}{\bibinfo{person}{Siyang Yuan}, \bibinfo{person}{Saurabh
  Gupta}, \bibinfo{person}{Xing Fan}, \bibinfo{person}{Derek Liu},
  \bibinfo{person}{Yang Liu}, {and} \bibinfo{person}{Chenlei Guo}.}
  \bibinfo{year}{2021}\natexlab{}.
\newblock \showarticletitle{Graph enhanced query rewriting for spoken language
  understanding system}. In \bibinfo{booktitle}{\emph{ICASSP 2021-2021 IEEE
  International Conference on Acoustics, Speech and Signal Processing
  (ICASSP)}}. IEEE, \bibinfo{pages}{7997--8001}.
\newblock


\end{thebibliography}

\appendix

\section{Rules for constraint user affinity}
\label{sec:appendix_x}

The following rules are applied to user affinity when building the collaborative user index as described in Section \ref{sec:affinity_based_personalized_index_expansion}
\begin{itemize}[itemsep=0pt,parsep=0pt,topsep=0pt,partopsep=0pt]
    \item We only consider entities such as songs, albums, artists, books, videos and shopping items for 3-hop affinity. In addition, there must be three unique 2-hop "user X→entity A→user Y" paths from user X to user Y to make sure there is a reasonable level of connections between them, otherwise we do not consider Y's affinity for expanding X's index. This index expansion increases coverage of user X's unseen defective interactions at both the entity level and the query level.
    \item We in addition consider entities such as genres, apps, cities, states, device/routine/contact names for only 2-hop affinity. These entities are intuitively less personalized and it is easier to introduce high-degree nodes along the paths with them. Therefore we use this constraint to make sure we only leverage rewrite candidates from the common affinity of X and Y for these types of entities. This expansion still increases query-level coverage of user X's unseen defective interactions even though we do not introduce new entities into X's index.
\end{itemize}

\section{Affinity Features \& Guardrail Features for Ranking}
\label{sec:appendix_y}

After retrieving potential rewrite candidates, the ranking module leverages a gradient boosting ranker model to select the most suitable rewrite. The current ranker has considered impression and various aforementioned defect signals as features. These features are calculated for  query utterances and entities at the global level and the user level. For instance, the user-level entity impression feature counts the number of times the entity appears in the user history. The same training data used for the retrieval model, but with the addition of features, is used for training the model.

We first introduce an additional level of features - \textit{multi-hop affinity features}; for example, the entity impression feature at the affinity level counts the occurrences of the entity in the user's constrained affinity as described Section \ref{sec:affinity_based_personalized_index_expansion}. In addition, we consider the following affinity features.
\begin{itemize}[itemsep=0pt,parsep=0pt,topsep=0pt,partopsep=0pt]
    \item The number of hops between the user and the rewrite candidate in the affinity, which can be 1,2,3 in our case.
    \item User X \& Y similarity features, such as the number of unique paths between the two users, the sum of impressions of these paths, the degree difference between X and Y, and the Jaccard distance between X and Y's neighborhood. All rewrites candidates along the "user X→...→user Y→..." paths have the same feature values for all user X \& Y similarity features.
\end{itemize}

We also introduce \textit{guardrail features} to counteract false trigger, and especially prevent the entity-swap error. A bad entity-swap is the typical type of error due to the enlarged index. One user queried “play songs by pink”, which indeed meant an artist named “Pink”. However, “Pink” was not present in the user history. Our enlarged index did cover the artist “Pink” associated with a rewrite candidate “play a million dreams by pink”, but it also introduced another artist “Pink Floyd” with a rewrite candidate “play songs from pink floyd”. The latter is falsely triggered due to its higher retrieval score from the current search-based QR system. We found query entity signals (e.g. impression, defect rate), and the similarities between query/rewrite entities are most critical to prevent the entity-swap error.

\end{document}